\documentclass[final,5p,times,twocolumn]{elsarticle}

\usepackage{amssymb}
\usepackage{amsmath}
\usepackage{amsthm}
\usepackage{amsfonts}

\usepackage{graphicx}
\usepackage{multirow}
\usepackage{booktabs}
\usepackage{soul}
\usepackage{color}
\usepackage{xcolor}
\usepackage{tcolorbox}
\usepackage{pifont}
\usepackage{colortbl}
\usepackage{makecell}
\usepackage{caption}
\usepackage{subcaption}
\usepackage{rotating}
\usepackage{bbm}
\usepackage{url}

\definecolor{}{RGB}{0, 255, 0}
\definecolor{blue}{RGB}{0, 0, 255}
\definecolor{red}{RGB}{255, 0, 0}
\definecolor{duong}{RGB}{0, 0, 255}
\definecolor{label1}{RGB}{149, 230, 255}

\newcommand{\vizsize}{0.20}
\newcommand{\textanglerot}{0}

\journal{Pattern Recognition Letters}

\begin{document}

\begin{frontmatter}



\title{PAT: Pixel-wise Adaptive Training for Long-tailed Segmentation}


\author[0]{Khoi Do} 
\author[1]{Minh-Duong Nguyen}
\author[2]{Nguyen H. Tran}
\author[3]{Viet Dung Nguyen\corref{cor1}}

\affiliation[0]{organization={Trinity College Dublin},
            addressline={College Green, Dublin 2}, 
            city={Dublin},
            postcode={}, 
            state={Dublin},
            country={Ireland}}
\affiliation[1]{organization={Pusan National University},
            addressline={2nd Busandaehak-ro 63beon-gil, Geumjeong-gu}, 
            city={Busan},
            postcode={}, 
            state={Busan},
            country={South Korea}}
\affiliation[1]{organization={The University of Sydney},
            addressline={Camperdown NSW 2050}, 
            city={NSW},
            postcode={}, 
            state={NSW},
            country={Australia}}
\affiliation[3]{organization={Hanoi University of Science and Technology},
            addressline={1st Dai Co Viet, Bach Khoa, Hai Ba Trung}, 
            city={Hanoi},
            postcode={}, 
            state={Hanoi},
            country={Vietnam}}

\begin{abstract}
Beyond class frequency, the impact of class-wise relationships among various class-specific predictions and the imbalance in label masks can cause significant problems in long-tailed segmentation learning. Addressing these challenges, we propose an innovative Pixel-wise Adaptive Training (PAT) technique tailored for long-tailed segmentation. PAT has two key features: 1) pixel-wise class-specific loss adaptation (PCLA), 2) head and tail balancing, and 3) low computation cost. First, PCLA tackles the detrimental impact of both rare classes within the long-tailed distribution and inaccurate predictions from previous training stages by \emph{encouraging learning classes with low prediction confidence}. Second, PAT integrates a new weighting curve function for \emph{guarding against forgetting classes with high confidence}. Third, PAT takes advantage of a pixel-wise weighting mechanism thus \emph{requiring computation cost just above Cross-Entropy}. PAT exhibits significant performance improvements, surpassing the current state-of-the-art by $2.2\%$ in the NyU dataset. Moreover, it enhances overall pixel-wise accuracy by $2.85\%$ and intersection over union value by $2.07\%$, with a particularly notable declination of $0.39\%$ in detecting rare classes compared to Balance Logits Variation, as demonstrated on the three popular datasets, i.e., OxfordPetIII, CityScape, and NyU. The code is available at \url{https://github.com/KhoiDOO/ibla}.
\end{abstract}


\begin{highlights}
\item Long-tailed rare objects cause unstable training Segmentation models.
\item Long-tailed datasets consist of imbalanced frequencies among masks and inside masks.
\item Adaptive weight from Class-sensitive learning loss function balance gradient learning.
\item Putting more weight on small objects while not forgetting high-confidence objects.
\item Low computation cost loss function for large-scale models and datasets.
\end{highlights}


\end{frontmatter}



\section{Introduction}\label{sec:intro}
\begin{figure*}[!htb]
    \centering
    \begin{subfigure}[b]{0.24\textwidth}
        \centering
        \includegraphics[width=0.9\textwidth]{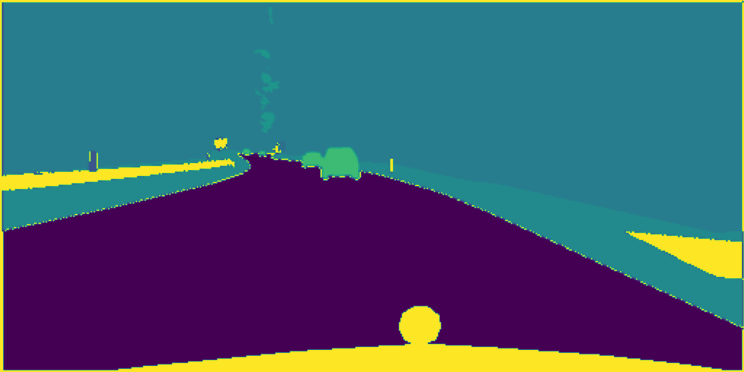}
        \includegraphics[width=\textwidth, height=0.6in]{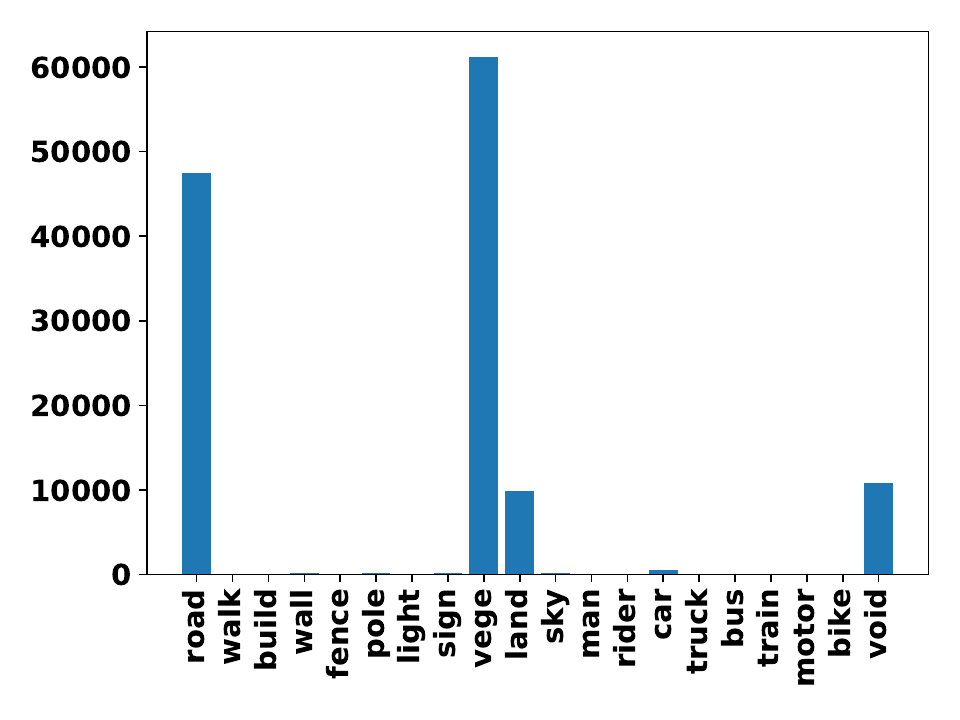}
        \caption{}
        \label{fig:a}
    \end{subfigure}
    \begin{subfigure}[b]{0.24\textwidth}
        \centering
        \includegraphics[width=0.9\textwidth]{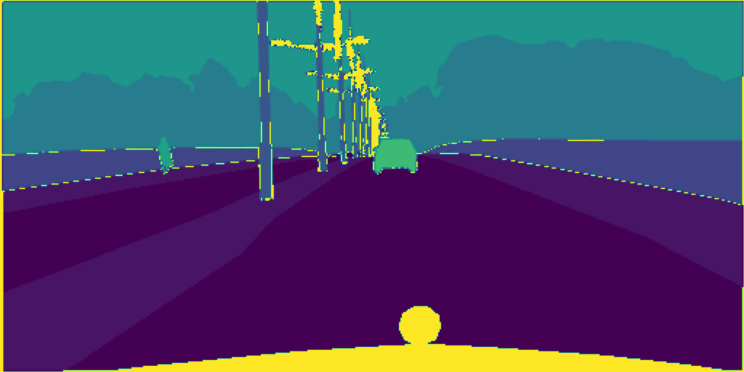}
        \includegraphics[width=\textwidth, height=0.6in]{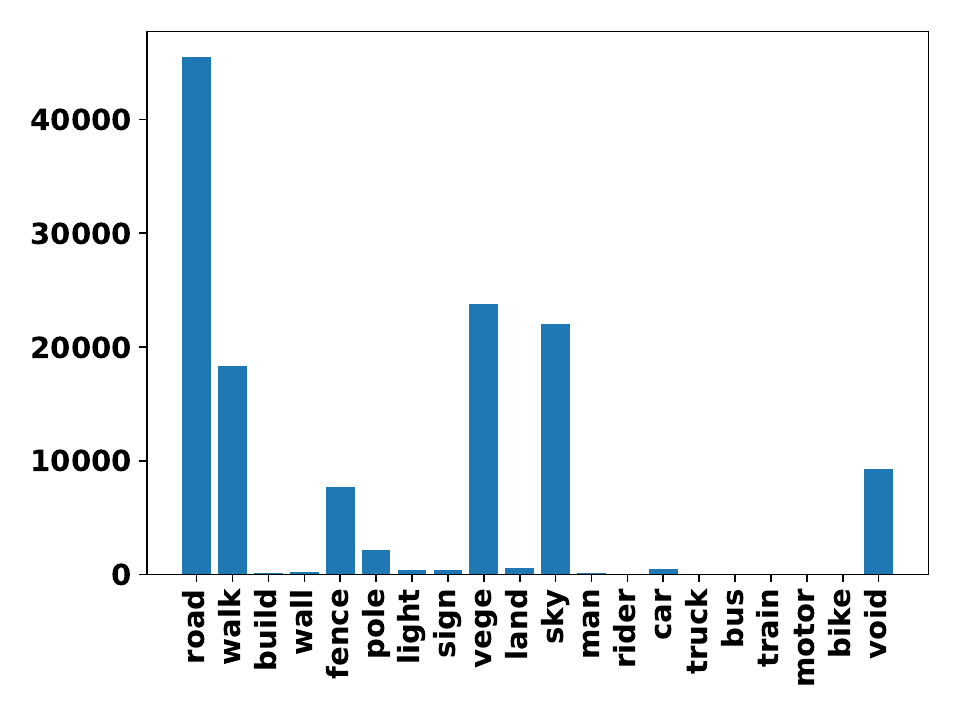}
        \caption{}
        \label{fig:b}
    \end{subfigure}
    \begin{subfigure}[b]{0.24\textwidth}
        \centering
        \includegraphics[width=0.9\textwidth]{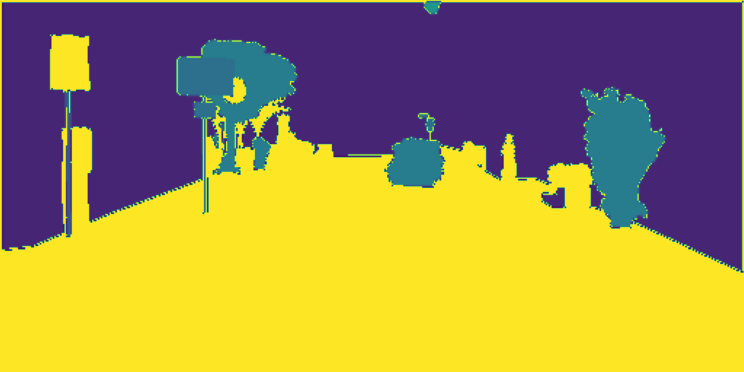}
        \includegraphics[width=\textwidth, height=0.6in]{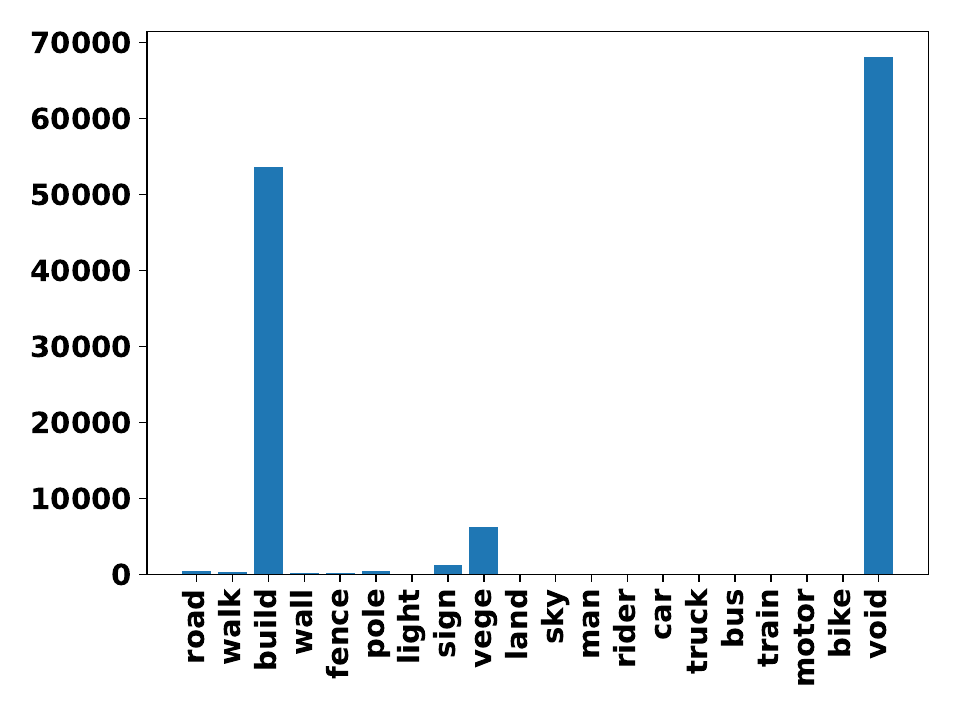}
        \caption{}
        \label{fig:c}
    \end{subfigure}
    \begin{subfigure}[b]{0.24\textwidth}
        \centering
        \includegraphics[width=0.9\textwidth]{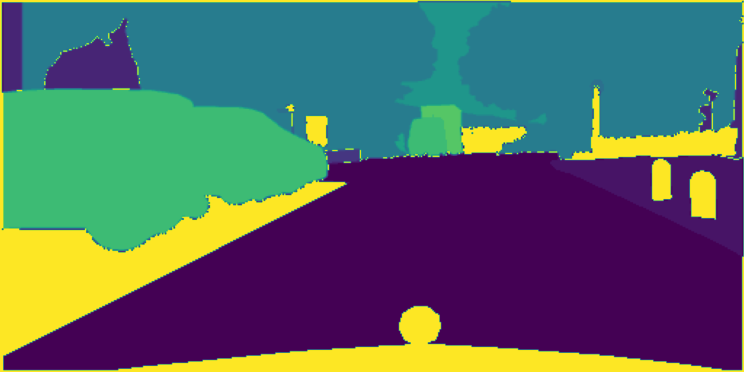}
        \includegraphics[width=\textwidth, height=0.6in]{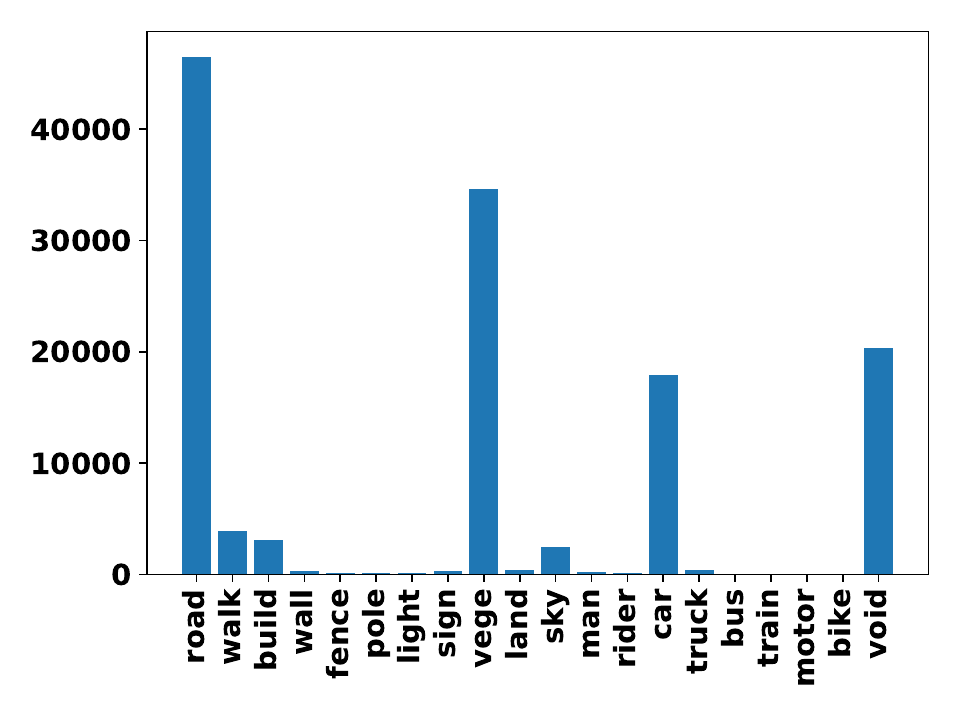}
        \caption{}
        \label{fig:d}
    \end{subfigure}
    \caption{Quantitative analysis on the imbalance in mask size among classes. The vertical axis illustrates the mask size calculated by the total number of pixels. The horizontal axis shows different masks that potentially appear in the ground truth. \ref{fig:a}) While road and vegetation masks take 50000 pixels and 60000 pixels, respectively, cars account for around 1000 pixels. 
    }
    \label{fig:imba-data-label}
\end{figure*} 
The applications of deep learning (DL) have shifted research interest toward datasets that are not perfectly balanced or meticulously crafted. It is crucial to explore robust algorithms that can perform well on imbalanced datasets. These challenges can be classified as distributional shifts between the training and testing sets\cite{2019-IBLA-LVIS}. Resampling \cite{jianhong-2023, bowen-2023}, data augmentation \cite{blv, perrett-2023}, logits adjustment (LA) \cite{he-2022, alexandridis-2022, wang-2021, li-2020, wang-2023}, and domain adaptation (DA) \cite{zhang-2022, li-2022, zang-2021} have been used to address long-tailed rare categories in segmentation. However, existing methods primarily focus on sample imbalance within classes, overlooking the issue of class imbalance within segmentation samples.

The drawbacks above originated from the following insights. \textbf{1) Imbalanced mask representations}: beyond the difficulties posed by rare objects, imbalanced mask representations occur when some masks dominate the learning process, leading to a bias towards recognizing dominant classes\cite{survey, blv, wang-2021}. \textbf{2) Model uncertainty and degradation}: models facing uncertainty often produce low-precision channel-wise logits, leading to biased gradient updates. These updates favor incorrect label predictions and ignore progress toward the true labels, further degrading performance\cite{focal, cb, ldam, bsl}. \textbf{3) High confidence categories neglection}: Current approaches focus on putting a higher weight on long-tailed rare objects while putting strictly small or zeroing weight causing unstable learning\cite{focal}. \textbf{4) High computation cost}: Although current approaches can tackle the issues\cite{ldam, blv, wang-2021}, they cost a large amount of computation.

We introduce Pixel-wise Adaptive Training (PAT), a novel approach for addressing long-tailed rare categories in segmentation. PAT comprises the following key contributions: \textbf{1) Pixel-wise Class-Specific Loss Adaptation (PCLA)}: This component focuses on pixel-wise predicting vectors (PPVs) within each pixel (see Fig.~\ref{fig:la}). By examining the PPVs, we can evaluate how individual channels influence learning by adjusting the logit predictions. \textbf{2) Head and Tail Balancing:} By employing an inverted softmax function, we can balance learning the presence of long-tailed rare objects and maintaining the model performance on high-confidence categories. \textbf{3) Low computation cost}: By taking full advantage of the Focal\cite{focal} weighting mechanism, we propose a new weighting curve that figures out long-tailed learning while requiring no supporting tensors, resulting in an extremely lower computation cost. 

\section{Related Works} 
Different from the traditional classification task, resampling methods\cite{jianhong-2023, bowen-2023} and data augmentation\cite{blv, perrett-2023, 9577506} cannot tackle the issues of long-tail distribution in segmentation as the ratio among class frequencies is not adjusted\cite{survey}. 

\textbf{Multi-stage \& Multi Objective Optimization}.
Multi-stage solution\cite{9577954, 9878867, 9879341, 9879371} consists of training feature extractor to learn representation of imbalance datasets, and fine-tuning classifier via frozen feature extractor. Otherwise, multi-objective optimization\cite{10203932, 10130611, 15555,10542424}, combining cross-entropy loss with contrastive loss or regularization.

\textbf{Logits adjustment}.
Logits adjustment (LA) is one of the most prominent techniques that can balance the effect of head class logits by equalizing the output logits distribution\cite{he-2022, alexandridis-2022, wang-2021, li-2020, wang-2023}. Domain adaptation\cite{zhang-2022, li-2022, zang-2021} is also considered to enhance the logits balancing, though accessing the target dataset is impractical in real-world applications\cite{domaingen-survey}. Post-processing is useful in removing the uncertainty\cite{pan-2021} at the pixel level in semantic segmentation, though it requires a calibration process for each dataset. 

\textbf{Class Sensitive Learning}.
Class-sensitive learning (CSL) is one direction in solving long-tailed learning\cite{survey}, which takes full advantage of classes' frequency to balance the logits distribution, class-wise gradient, etc. LA and CSL are the two most easy-to-use yet effective methods in long-tailed learning. Focal (Focal) function\cite{focal} is firstly proposed to tackle this challenge by taking the inverse of logits as a weight for each class. In\cite{cb}, a class-balance loss function is proposed, working as an effective sampler by a class frequency weighting function. The combination of class-balance loss and Focal is also considered in\cite{cb}. Balancing the effect of the exponential function in the Softmax activation function is studied\cite{wang-2021, li-2020}. Other recent methods\cite{ldam, blv} focus on using noise in logits to balance them which are also popular in segmentation, we also compare our proposed method with those. We also show that our proposed method surpasses the others in both performance and utilization. 

\section{Proposed Method}
\begin{figure*}
    \centering
    \includegraphics[width=0.7\textwidth]{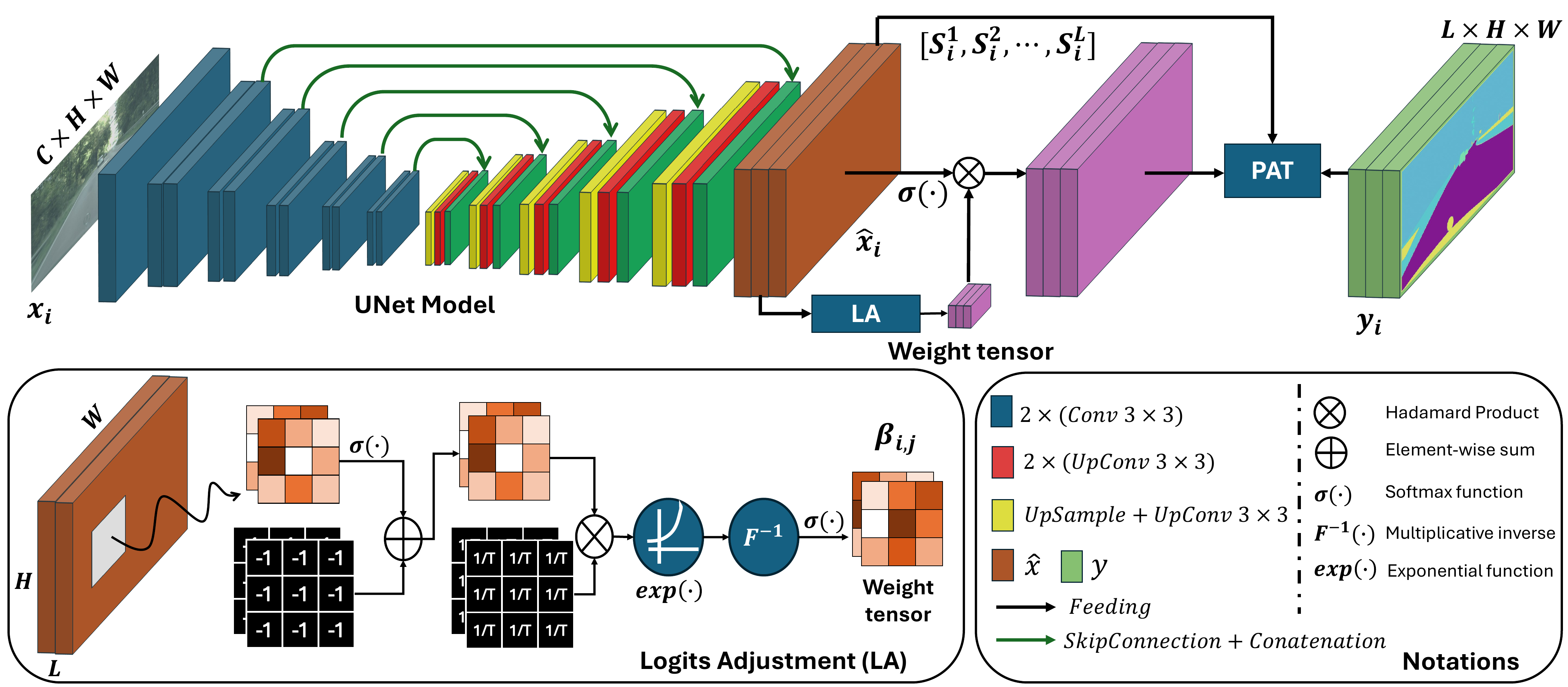}
    \caption{Overall methodology. \textbf{1) Training procedure}: an image $x_i$ is fed into an encoder-decoder architecture to produce logits $\hat{x}_i$. Subsequently, $\hat{x}_i$ is adjusted to create a weight tensor, which has the same size as $\hat{x}_i$. The normalized $\hat{x}_i$ is multiplied by the weight tensor to equalize the contribution of each logit to the PAT loss function.
    \textbf{2) Logits Adjustment}: Logits vector is normalized by the Softmax, added by a tensor of $-1$, and scaled by the exponential function to find the inverse dominant coefficients $\beta_{i, j}$. Then, $\beta_{i,j}$ is normalized to $[0, 1]$ to form the weight tensor.}
    \label{fig:swat_arch}
\end{figure*}
\subsection{Problem Statement}
Consider $(x,y) \sim P(\mathcal{X}, \mathcal{Y})$, where $x\in\mathbb{R}^{N\times C\times H\times W}$ and $y\in\mathbb{R}^{N\times L\times H\times W}$ are input data and corresponding ground truth, respectively. $C$ and $L$ are the number of image channels and categories, respectively. $N$, $H$, and $W$ are the total number of training samples, height, and width of the image, separately. The segmentation problem is represented in Eq.~(\ref{eq:over}).
\begin{flalign}\label{eq:over}
    \mathcal{L}(x, y) &= \frac{1}{B}\sum^{B-1}_{i=0}\sum^{L-1}_{l=0}\mathcal{L}_l(x_i, y_i),&&
\end{flalign}
where $\mathcal{L}_l(x_i, y_i) = -\log\Big(\frac{\exp\{\hat{x}^l_{i}\}}{\sum_{l'=0}^{L-1}\exp\{\hat{x}^{l'}_{i}\}}\Big) y^l_{i}$ denotes the class-wise loss on sample $x_i\in\mathbb{R}^{C\times H\times W}$ and its corresponding ground truth $y_i\in\mathbb{R}^{L\times H\times W}$. $\hat{x}^l_{i}, y^l_i\in\mathbb{R}^{H\times W}$ are the predicted mask and the ground-truth of channel $l$ (which represents class $l$), respectively.
\subsection{Imbalance among label masks}\label{sec:loss-dec}
One major challenge in image segmentation is the class imbalance in label masks. Larger masks contribute more significantly to the loss of function than smaller masks, leading to a bias towards dominant classes. Specifically, we come over the class-wise loss component, which can be represented as:
\begin{flalign}\label{eq:ibla-label-mask}
    \mathcal{L}_l(x_i, y_i)
    &= -\sum^{HW}_{j=0} \log\Big(\frac{\exp\{\hat{x}^l_{i,j}\}}{\sum_{l'=0}^{L-1}\exp\{\hat{x}^{l'}_{i,j}\}}\Big) y^l_{i,j} &&\\ \nonumber
    &= -\sum^{HW}_{j=0} \log\Big(\frac{\exp\{\hat{x}^l_{i,j}\}}{\sum_{l'=0}^{L-1}\exp\{\hat{x}^{l'}_{i,j}\}}\Big) \mathbbm{I}(y^l_{i,j}=1) = S^l_i \times \Bar{\ell}_l (x_i,y_i), &&\\ \nonumber
\end{flalign}
where $S^l_i$ and $\Bar{\ell}_l (x_i,y_i)$ denote the size of label $l$ mask and the cross-entropy value on class $l$ for image $i$, respectively, where $S^l_i = \sum^{HW}_{j=0} \mathbbm{1}(y^l_{i,j} = 1)$.

While the traditional approach is rooted in classification problems, in segmentation tasks, the loss is adjusted based on the mask size $S^l_i$. Consequently, to ensure uniformity in gradient magnitude, we diminish the loss by the label mask size of each instance. This adjustment guarantees that all class-specific loss pixels receive equal consideration within the collective loss function.
\subsection{Pixel-wise Adaptive Traning with Loss Scaling}\label{sec:pat}
The summary of the methodology is shown in Fig. \ref{fig:swat_arch}. To design an adaptive pixel-wise loss scaling, we first decompose the conventional segmentation function into a pixel-wise function (refer to Eq. (\ref{eq:over2})), which is derived from Eq. (\ref{eq:over}).
\begin{flalign}\label{eq:over2}
    &\mathcal{L}(x, y) = -\frac{1}{B}\sum^{B - 1}_{i=0} \sum^{HW}_{j=1} 
    \Biggr[\sum^{L-1}_{l=0}y^l_{i,j}\log\Big(\frac{\exp\{\hat{x}^l_{i,j}\}}{\sum_{l'=0}^{L-1}\exp\{\hat{x}^{l'}_{i,j}\}}\Big) \Biggr],&&
\end{flalign}
where $\hat{x}^l_{i, j}$ is the logits prediction of sample $i$ at pixel $j$ with regard to category $l$. 
In segmentation problems, the class is considered by a composition of many pixels over the masks. We hypothesize that the learning in each class may occur diversely according to the classification of different pixels. Therefore, we propose the pixel-wise adaptive (PAT) loss via the pixel-wise adaptive coefficient set $\beta_{i,j}\in\mathbb{R}^L = [\beta_{i,j}^0, \beta_{i,j}^1, \cdots, \beta_{i,j}^{L-1}]$. 
\begin{flalign}\label{eq:over3}
    \mathcal{L}(x, y) = -\frac{1}{B}\sum^{B - 1}_{i=0} \sum^{HW}_{j=1}\sum^{L-1}_{l=0} \beta_{i,j}^l \times \frac{y^l_{i,j}}{S^l_i}\log\Big(\frac{\exp\{\hat{x}^l_{i,j}\}}{\sum_{l'=0}^{L-1}\exp\{\hat{x}^{l'}_{i,j}\}}\Big).&&
\end{flalign}
\begin{figure*}[!htb]
     \centering
     \begin{subfigure}[b]{0.325\textwidth}
         \centering
         \includegraphics[width=\textwidth]{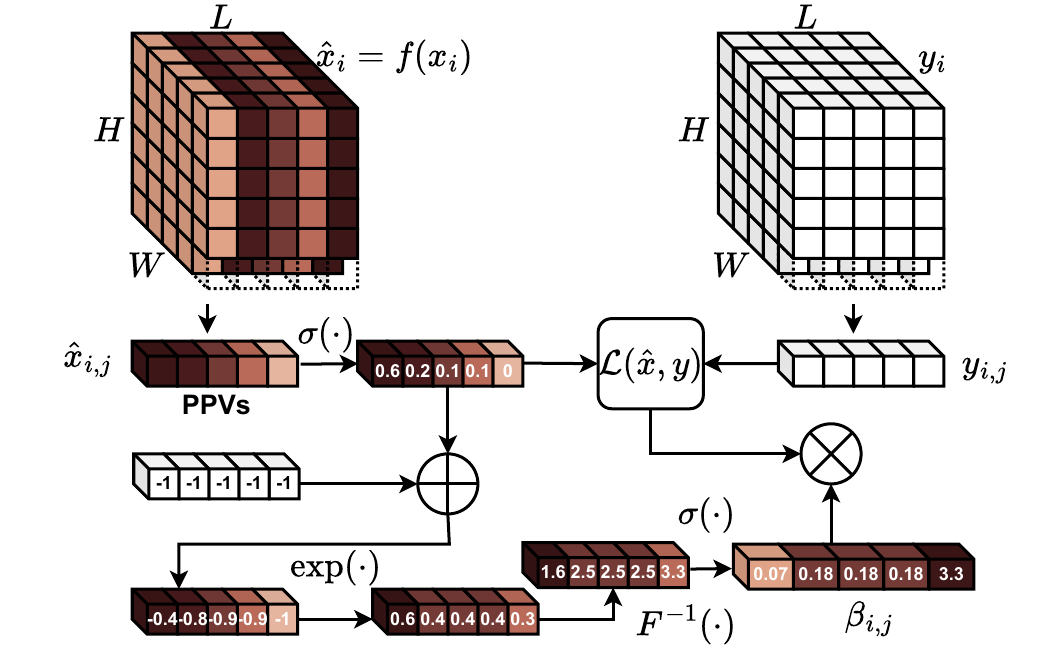}
         \caption{}
         \label{fig:la}
     \end{subfigure}
     \begin{subfigure}[b]{0.325\textwidth}
         \centering
         \includegraphics[width=\textwidth]{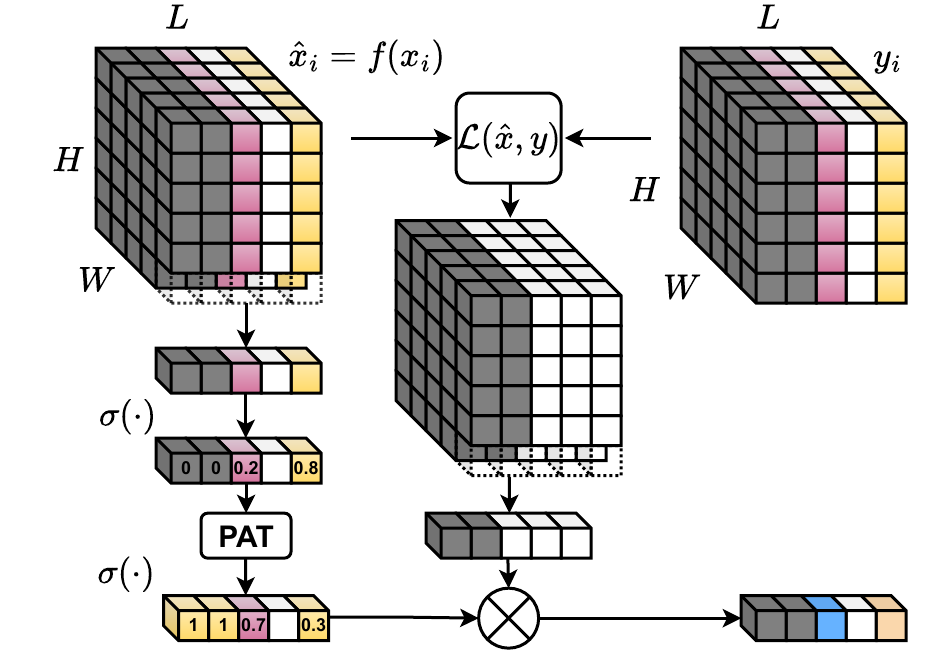}
         \caption{}
         \label{fig:nullgrad}
     \end{subfigure}
     \begin{subfigure}[b]{0.325\textwidth}
         \centering
         \includegraphics[width=\textwidth]{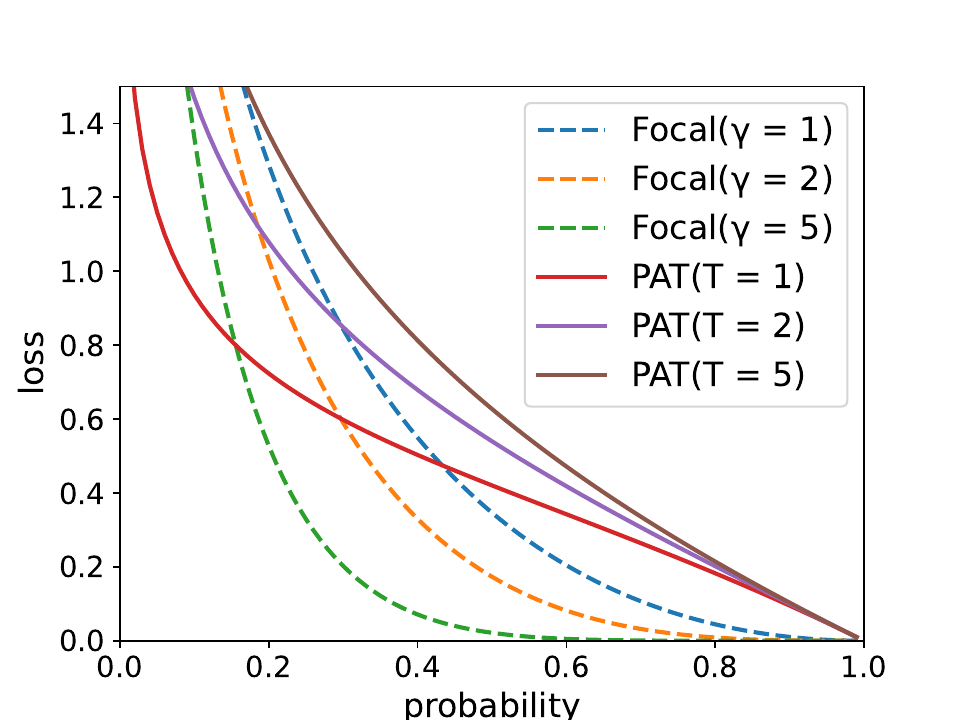}
         \caption{}
         \label{fig:vs}
     \end{subfigure}
     \caption{Fig. \ref{fig:la} illustrates the PAT procedure of adjusting the logits' value to tackle the imbalance in dominant probability from categories whose big mask size. Fig. \ref{fig:nullgrad} shows the process of adaptive gradient scaling in PAT. Specifically, the channels with no mask can easily be adapted. Therefore, the problem of adaptive gradient scaling can be reduced to two cases. In addition to Fig. \ref{fig:la}, Fig. \ref{fig:vs} shows the difference in scaling coefficient between PAT and Focal\cite{focal}, that PAT (smooth lines) (i) puts a higher weight on low confidence pixel and (ii) keeps low scaling coefficients for high confidence pixels. Otherwise, Focal (dash line), puts zero scalarization on well-classified pixels that may cause forgetfulness of frequent or big mask size categories.}
     \label{fig:ladaptiveGradientScaling}
\end{figure*}

Our key idea is to control the PAT loss using $\beta_{i,j}$. Essentially, $\beta_{i,j}$ represents a tensor with dimensions identical to the logits $\hat{x}_{i,j}$. Through the pixel-wise multiplication, $\beta_{i,j}$ effectively modulates the pixel-wise loss components. Calculation of $\beta_{i,j}$ based on the logits $\hat{x}_{i,j}$ is as follows:
\begin{flalign}
    \beta_{i,j} = \frac{1}{\exp\left\{(p(\hat{x}_{i,j}) - 1 + \epsilon) /T\right\}},&&
\label{eq:PAT}
\end{flalign}
where $p(\hat{x}_{i,j}) = \left[\exp\{\hat{x}^l_{i, j}\}/\sum_{l'=0}^{L-1}\exp\{\hat{x}^{l'}_{i,j}\}\right]_{l=0}^{L-1}$ indicates the output of the softmax activation which normalize the logits vector elements into probability range of $[0, 1)$. We define $T$ as a temperature coefficient and $\epsilon$ as an arbitrary constant. By tuning the $\beta_{i,j}$ according to each pixel-wise vector $\hat{x}_{i, j} = \{\hat{x}^l_{i, j}\vert ~l\in \{0,\ldots, L-1\}\}$, our proposed coefficient hinges on two key concepts: firstly, equalizing the loss value across various logits, and secondly, preventing negative transfer in well-classified results. Further analysis is presented in Section~\ref{sec:logit-imbalance}. Additionally, in Eq.~\eqref{eq:over3}, we normalize the loss across all components by dividing by $S^l_i$. This approach allows us to penalize loss components that have a dominant size relative to others, thereby facilitating the class-wise gradient magnitudes.

\section{Empirical Analysis}
In Section~\ref{sec:generalization-PAT}, we demonstrate that our proposed PAT effectively addresses the numerical instability that may arise due to Eq.~\eqref{eq:PAT}. In Section~\ref{sec:logit-imbalance}, we prove that PAT can effectively handle long-tailed rare object segmentation, particularly in detecting objects with small portions of the mask. Moreover, PAT maintains the performance of high-confidence classes. In Section~\ref{sec:space-time}, we show that PAT achieves low computational costs compared to existing state-of-the-art methods by incorporating a simple weighting mechanism that does not require additional supporting tensors or calculations.

\subsection{Generalization of PAT on special case}\label{sec:generalization-PAT}
In numerous scenarios, $\beta_{i,j}$ may encounter near-zero logit values, potentially causing value explosions. This occurrence can lead to computational errors in practice. To mitigate this issue, we introduce temperature coefficients $T$ and a constant $\epsilon$, effectively preventing the value explosion of $\beta_{i,j}$. Moreover, near-zero logit values are often associated with the absence of label masks. Consequently, through the computation of the joint loss function, pixel-level loss values are frequently nullified to $0$ rather than undergoing explosion (see Fig.~\ref{fig:nullgrad}).

\subsection{Analysis on the PAT to the logits imbalance}\label{sec:logit-imbalance}
Experimentally, Focal loss performance is lower than current approaches, through its simple and optimized implementation. To have a comprehensive understanding of PAT robustness to the imbalance rare object segmentations, we compare the loss value of PAT and Focal\cite{focal} at different logit probabilities in Fig.~\ref{fig:vs}, yielding two significant observations. 
\begingroup
\setlength{\tabcolsep}{7pt}
\begin{table}[!htb]
\centering
\caption{Adaptive loss value comparison between Focal loss and PAT loss functions with different adaptive coefficients: $\gamma\in\{2, 5\}$ and $T\in\{2, 5\}$.} 
\resizebox{0.3\textwidth}{!}{%
\begin{tabular}{@{}cccccc@{}}
\toprule
$\boldsymbol{p(\hat{x}_{i, j})}$  & 0.2  & 0.3  & 0.4  & 0.5 \\ \midrule
\textbf{Focal} ($\gamma = 2$) & 1.03 & 0.59 & 0.33 & 0.17  \\ \midrule
\textbf{PAT} ($T=2$) & 1.08 & 0.85 & 0.68 & 0.54 \\ \midrule
\textbf{Focal} ($\gamma = 5$) & 0.53 & 0.2  & 0.07 & 0.02 \\ \midrule
\textbf{PAT} ($T=5$) & 1.37 & 1.05 & 0.81 & 0.63 \\ \bottomrule
$\boldsymbol{p(\hat{x}_{i, j})}$ & 0.6  & 0.7  & 0.8  & 0.9 \\ \midrule
\textbf{Focal} ($\gamma = 2$) & \textbf{0.08} & \textbf{0.03} & \textbf{0.01} & \textbf{0.0} \\ \midrule
\textbf{PAT} ($T=2$) & 0.42 & 0.31 & 0.2  & 0.1 \\ \midrule
\textbf{Focal} ($\gamma = 5$) & \textbf{0.01} & \textbf{0.0}  & \textbf{0.0}  & \textbf{0.0} \\ \midrule
\textbf{PAT} ($T=5$) & 0.47 & 0.34 & 0.21 & 0.1 \\ \bottomrule
\end{tabular}}
\label{tab:logit-imblance}
\end{table}
\endgroup

Firstly, by parameterizing the loss function with the PAT scaling factor, we observe more balanced learning across classes with varying logit probabilities. Secondly, in contrast to Focal, we notice that losses associated with high-probability logits are zero-weighted (refers to Tab. \ref{tab:logit-imblance}). This phenomenon can prevent positive transfer on well-classified samples. In comparison, the PAT-scaling parameterized loss function fosters equitable learning while maintaining loss information for high-probability logits, thus enabling stable training.

\subsection{Analysis on time and space complexity}\label{sec:space-time}
As demand for low-cost computation algorithm\cite{survey, dlv3, dlv3p}, we take into consideration time and space complexity. Tab. \ref{tab:complexity} theoretically suggests that traditional Cross-Entropy, Focal, and PAT have the lowest complexity in both time and space, compared to LDAM and BLV, which require extensive supporting tensors ($\Delta_y$, $\delta(\sigma)$, and $c$) to manipulate margin probability distribution.

\begingroup
\setlength{\tabcolsep}{5pt}
\begin{table}[!ht]
\centering
\caption{Theoretical time ($\mathcal{O}$) and space ($\mathcal{V}$) complexity comparison between state-of-the-arts and PAT. Denote $\mathcal{F}$ as loss formula, $\gamma$ is scale coefficient of Focal\cite{focal}, $z_y$, $z_i$ are logits, $\sigma$ is standard deviation, $\delta$ is a distribution generator, $\psi = BCHW$, and $\Bar{p}(x) = 1 - p(x)$}
\begin{tabular}{@{}p{1cm}ccc@{}}
\toprule
\textbf{Method} & $\boldsymbol{\mathcal{F}}$ & $\boldsymbol{\mathcal{O}}$ & $\boldsymbol{\mathcal{V}}$ \\ \midrule
CE     & $-\log(p(x))$ & $\mathcal{O}(\psi)$ & $\mathcal{O}(\psi)$ \\ \midrule
Focal  & $-(\Bar{p}(x))^\gamma\log(p(x))$ & $\mathcal{O}(\psi)$ & $\mathcal{O}(\psi)$ \\ \midrule
LDAM   & $-\log(\frac{\exp(z_y - \Delta_y)}{\sum_i\exp(z_i - \Delta_i)})$ & $\mathcal{O}(2\psi)$ & $\mathcal{O}(2\psi) + \mathcal{O}(2C)$ \\ \midrule
BLV    & $-\log(\frac{\exp(z_y + c_y\delta(\sigma))}{\sum_i\exp(z_i + c_i\delta(\sigma))})$  & $\mathcal{O}(2\psi)$ & $\mathcal{O}(3\psi) + \mathcal{O}(2C)$ \\ \midrule
PAT    & $-\exp\{(\Bar{p}(x))/T\}\log(p(x))$ & $\mathcal{O}(\psi)$ & $\mathcal{O}(\psi)$ \\ \bottomrule
\end{tabular}
\label{tab:complexity}
\end{table}
\endgroup

\section{Experimental Evaluations and Discussion}\label{sec:evaluation}
\begin{table*}[!ht]
\caption{Overall Performance of 8 baselines (i.e. Vanilla Softmax, Focal, Class Balance Loss, Class Balance Focal) and proposed method among three different scenarios including OxfordPetIII, CityScape, and NyU datasets. In detail, the bold number indicates the best performance, while $\uparrow$ and $\downarrow$ show "higher is better" and "lower is better", respectively. Note that all experiments shown in this table are trained using SegNet\cite{segnet} architecture.}
\centering
\resizebox{0.8\textwidth}{!}{%
\begin{tabular}{p{3.5cm}ccccccccc}
\toprule
\multicolumn{1}{c}{\multirow{3}{*}{\textbf{Method}}} &
  \multicolumn{9}{c}{\textbf{Dataset}} \\ \cline{2-10} 
\multicolumn{1}{c}{} &
  \multicolumn{3}{c}{\textbf{OxfordPetIII\cite{oxfordpet}}} &
  \multicolumn{3}{c}{\textbf{CityScape\cite{city}}} &
  \multicolumn{3}{c}{\textbf{NyU\cite{nyu}}} \\
\multicolumn{1}{c}{} &
  \multicolumn{1}{l}{mIoU$\uparrow$} &
  \multicolumn{1}{l}{Pix Acc$\uparrow$} &
  \multicolumn{1}{l}{Dice Err$\downarrow$} &
  \multicolumn{1}{l}{mIoU$\uparrow$} &
  \multicolumn{1}{l}{Pix Acc$\uparrow$} &
  \multicolumn{1}{l}{Dice Err$\downarrow$} &
  \multicolumn{1}{l}{mIoU$\uparrow$} &
  \multicolumn{1}{l}{Pix Acc$\uparrow$} &
  \multicolumn{1}{l}{Dice Err$\downarrow$} \\ \midrule
CE &
  76.14 &
  91.21 &
  \textbf{0.23} &
  73.83 &
  85.31 &
  0.66 &
  18.05 &
  53.50 &
  1.80 \\
\makecell[l]{Focal ($\gamma=2$)} &
  75.76 &
  91.17 &
  0.30 &
  74.02 &
  85.44 &
  0.56 &
  15.14 &
  51.07 &
  1.75 \\
CB &
  76.60 &
  90.90 &
  \vizsize &
  72.26 &
  81.33 &
  0.69 &
  18.56 &
  52.17 &
  1.94 \\
\makecell[l]{CBFocal ($\gamma=2$)} &
  76.02 &
  90.54 &
  0.26 &
  71.17 &
  80.70 &
  0.72 &
  17.31 &
  50.46 &
  1.76 \\
BMS &
  13.22 &
  25.45 &
  1.28 &
  8.15 &
  11.80 &
  3.08 &
  12.27 &
  22.84 &
  2.40 \\
\makecell[l]{LDAM ($\mu=0.5$, $s = 20$)}&
  75.43 &
  90.97 &
  0.78 &
  74.80 &
  85.20 &
  2.27 &
  19.59 &
  52.59 &
  2.30 \\
\makecell[l]{BLV (Gaussian, $\sigma = 0.5$)} &
  76.24 &
  91.22 &
  \textbf{0.23} &
  74.21 &
  85.37 &
  0.53 &
  18.37 &
  52.62 &
  1.90 \\ \midrule
\makecell[l]{PAT (Ours) ($T = 20$)} &
  \textbf{\begin{tabular}[c]{@{}c@{}}76.69 \textcolor{}{$\uparrow$ 0.09}\end{tabular}} &
  \textbf{\begin{tabular}[c]{@{}c@{}}91.28 \textcolor{}{$\uparrow$ 0.07}\end{tabular}} &
  \textbf{0.23} &
  \textbf{\begin{tabular}[c]{@{}c@{}}76.22 \textcolor{}{$\uparrow$ 2.2}\end{tabular}} &
  \textbf{\begin{tabular}[c]{@{}c@{}}85.80 \textcolor{}{$\uparrow$ 0.36}\end{tabular}} &
  \textbf{\begin{tabular}[c]{@{}c@{}}0.51 \textcolor{black}{$\downarrow$ 0.02}\end{tabular}} &
  \textbf{\begin{tabular}[c]{@{}c@{}}21.41 \textcolor{}{$\uparrow$ 2.85}\end{tabular}} &
  \textbf{\begin{tabular}[c]{@{}c@{}}55.57 \textcolor{}{$\uparrow$ 2.07}\end{tabular}} &
  \textbf{\begin{tabular}[c]{@{}c@{}}1.36 \textcolor{blue}{$\downarrow$ 0.39}\end{tabular}} \\\bottomrule
\end{tabular}}
\label{tab:over}
\end{table*}
We conducted experiments on three popular datasets: OxfordPet\cite{oxfordpet}, CityScapes\cite{city}, and NyU\cite{nyu} whose frequency of classes is considerably sample-wise imbalanced (refers to Section \ref{sec:sota}). The training, validating, and testing ratios are 0.8, 0.1, and 0.1, respectively. OxfordPetIII\cite{oxfordpet} Dataset contains 37 dog/cat categories ($\approx$200 images/category). Mask ground truth has 3 classes: background, boundary, and main body. Images are original $640\times40$, resized to $256\times256$. The biggest obstacle is segmenting the boundary of the animal which is very small and not easily distinguishable. CityScapes\cite{city} dataset contains roughly 5000 images with a high resolution of $1024\times512$ pixels. This dataset faced a big issue in a high number of imbalanced class imbalances (refers to Fig. \ref{fig:imba-data-label}). The NyU\cite{nyu} dataset contains roughly 1000 samples whose size of $340\times256$ pixels. As in the CityScapes dataset, NyU also faces a big challenge in segmenting long-tailed rare objects.

We compare PAT with Focal\cite{focal}, Class Balance Loss (CB)\cite{cb}, the combination of CB and Focal (CBFocal)\cite{cb}, Balance Meta Softmax (BMS)\cite{bsl}, Label Distribution Aware Margin Loss (LDAM)\cite{alexandridis-2022}, and Balance Logits Variation (BLV)\cite{blv} evaluated by mean Intersection over Union (mIoU \%), pixel accuracy (Pix Acc \%), and Dice Error (Dice Err). The number of rounds of all experiments is fixed to 30000 rounds. 

We tuned the hyperparameter of each loss function to find the best case. Specifically, 1) Focal and the combination of Class Balance and Focal are trained with different values of $\gamma\in\{0.5, 1, 2, 3, 4, 5\}$\cite{focal,cb}. 2) In the LDAM, we set the parameter $\mu$ of $0.5$ as the default setting in\cite{ldam} and trained with different scale $s\in\{10, 20, 30, 50\}$. 3) For the BLV, we applied types of distribution: Gaussian, Uniform, and Xaviver along with different standard deviation $\sigma\in\{0.5, 1, 2\}$. 4) Hyper-parameter tuning is conducted on PAT with values of $T\in\{5, 10, 20, 50\}$ (refer to Section \ref{sec:abs-temp}). 

\subsection{Comparisons to state-of-the-arts (SOTA)}\label{sec:sota}
\begin{figure*}[!ht]
    \centering
    \resizebox{0.7\textwidth}{!}{%
    \begin{tabular}{lcccc}
       \rotatebox{\textanglerot}{Image} & \includegraphics[width=\vizsize\textwidth]{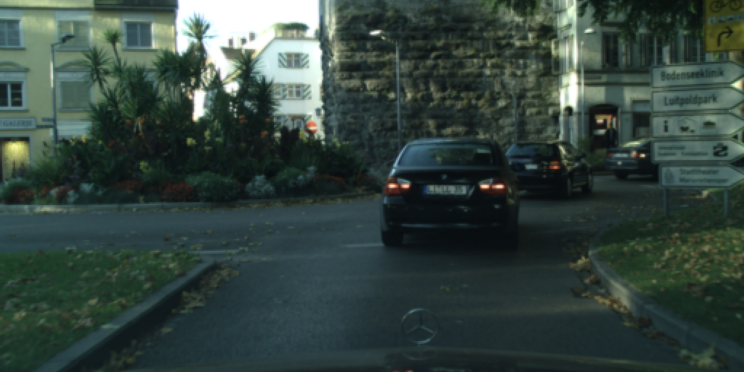} & \includegraphics[width=\vizsize\textwidth]{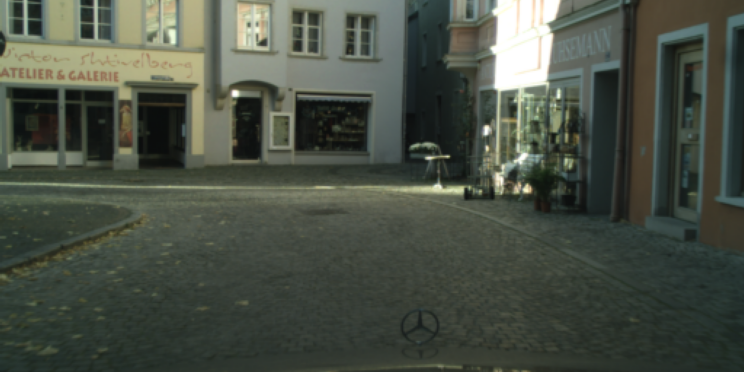}
       & \includegraphics[width=\vizsize\textwidth]{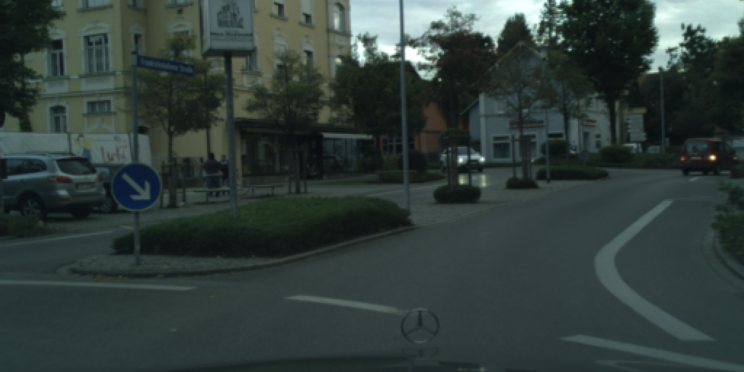} 
       & \includegraphics[width=\vizsize\textwidth]{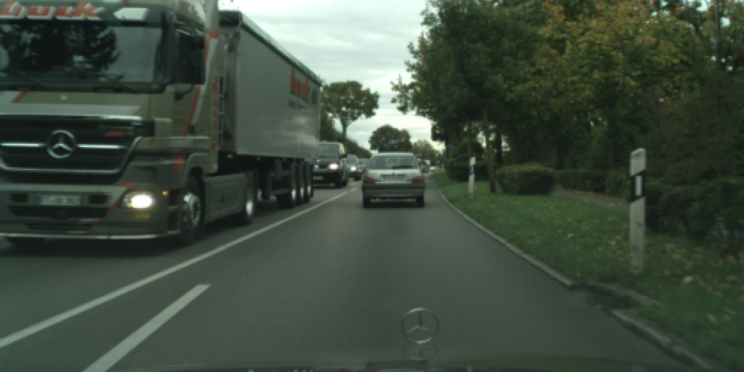}\\
       \rotatebox{\textanglerot}{Label} & \includegraphics[width=\vizsize\textwidth]{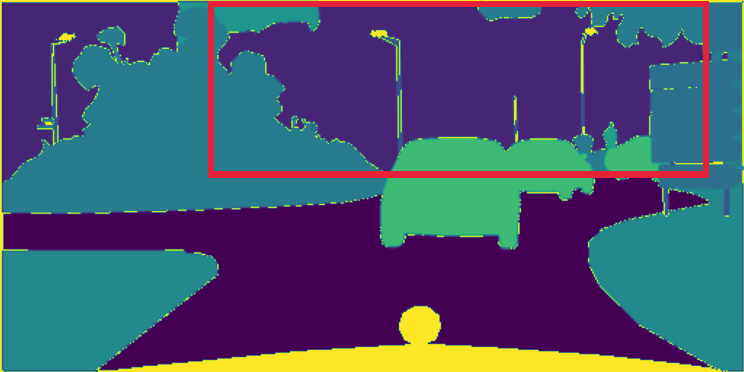} & \includegraphics[width=\vizsize\textwidth]{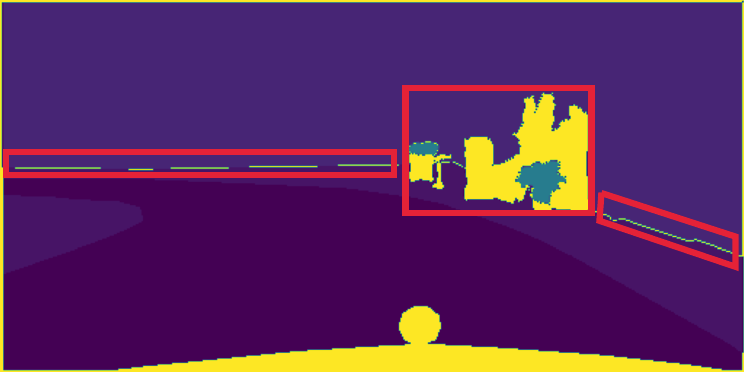}
       & \includegraphics[width=\vizsize\textwidth]{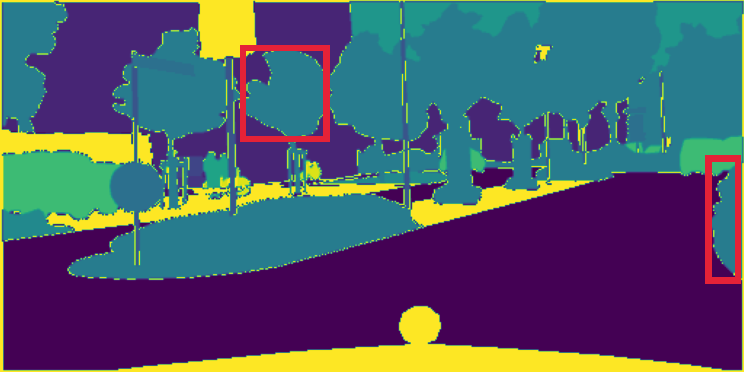} 
       & \includegraphics[width=\vizsize\textwidth]{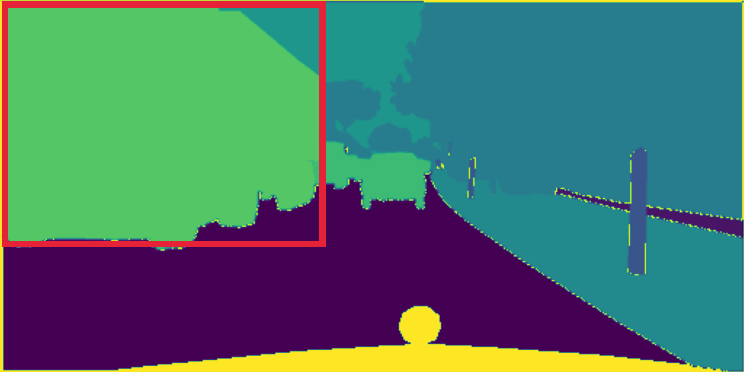}\\
       \rotatebox{\textanglerot}{PAT (Ours)} & \includegraphics[width=\vizsize\textwidth]{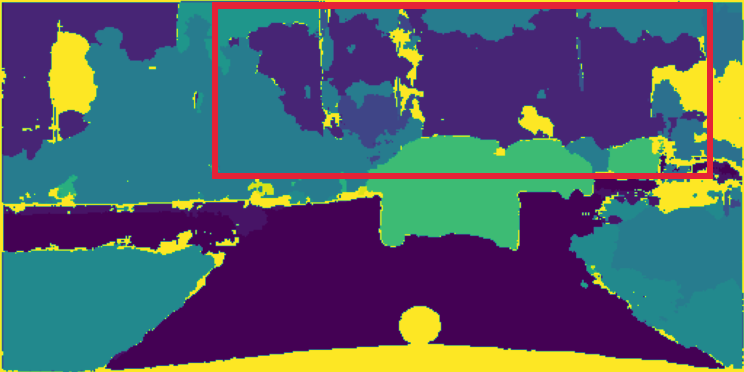} & \includegraphics[width=\vizsize\textwidth]{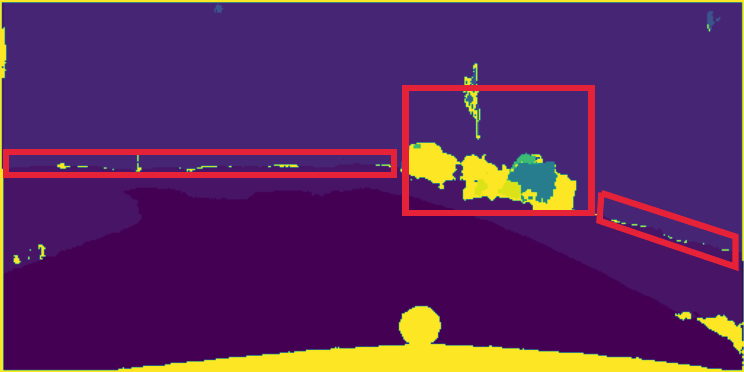}
       & \includegraphics[width=\vizsize\textwidth]{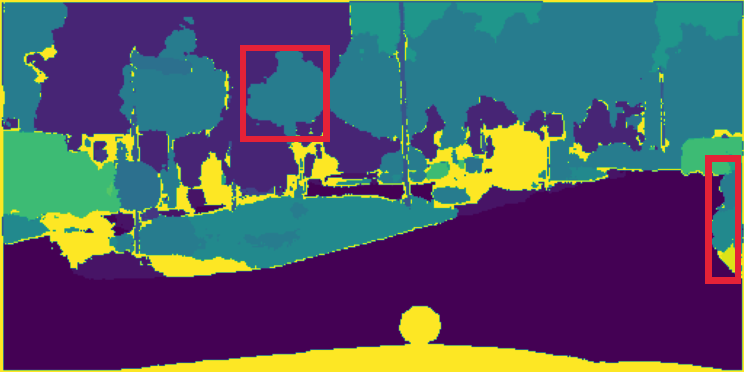} 
       & \includegraphics[width=\vizsize\textwidth]{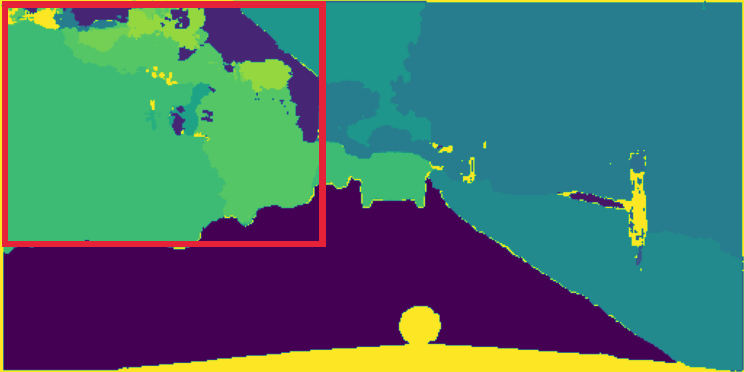}\\
       \rotatebox{\textanglerot}{Focal} & \includegraphics[width=\vizsize\textwidth]{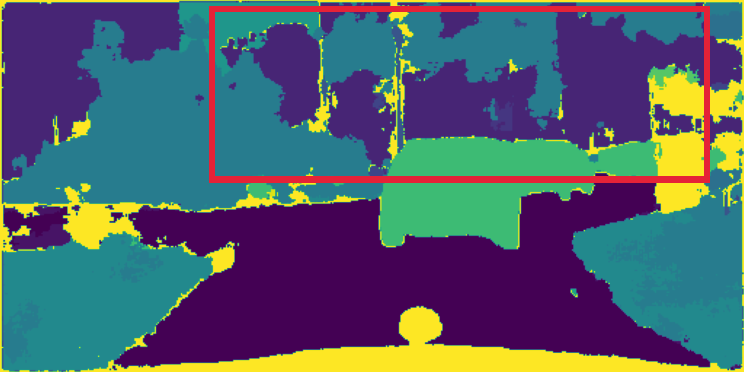} & \includegraphics[width=\vizsize\textwidth]{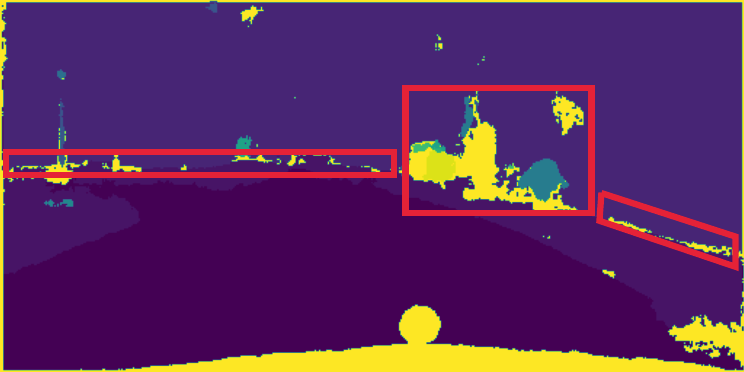}
       & \includegraphics[width=\vizsize\textwidth]{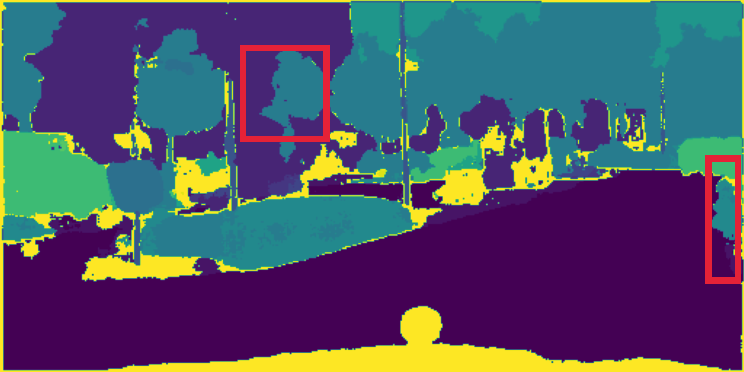} 
       & \includegraphics[width=\vizsize\textwidth]{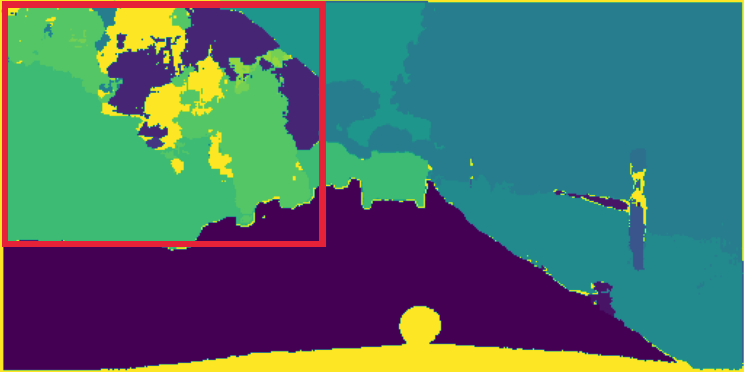}\\
       \rotatebox{\textanglerot}{CB} & \includegraphics[width=\vizsize\textwidth]{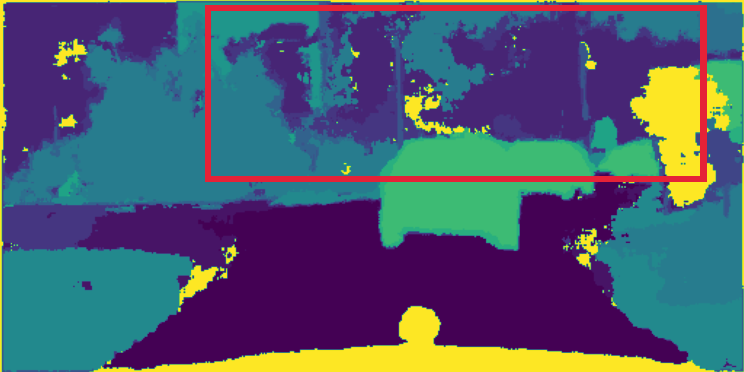} & \includegraphics[width=\vizsize\textwidth]{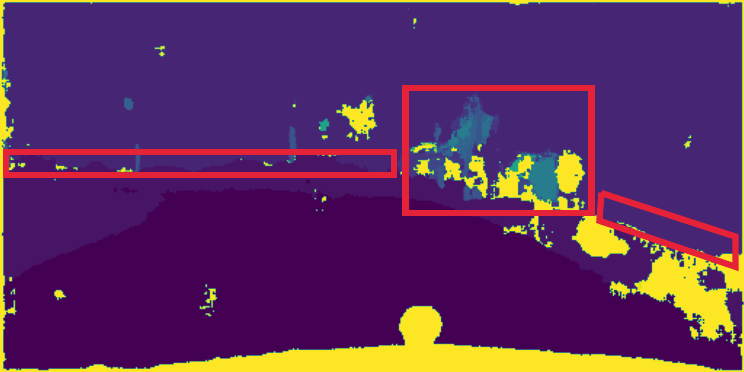}
       & \includegraphics[width=\vizsize\textwidth]{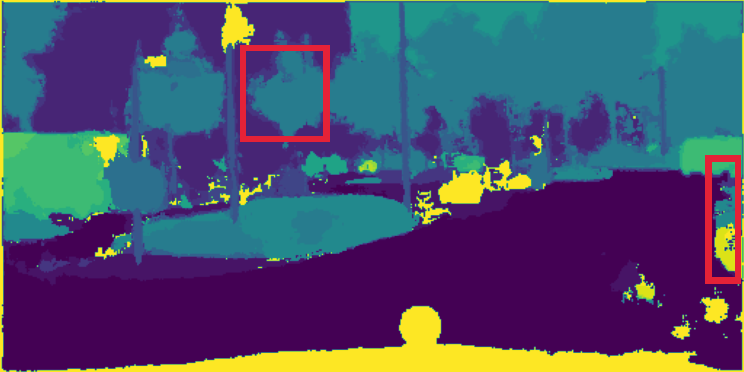} 
       & \includegraphics[width=\vizsize\textwidth]{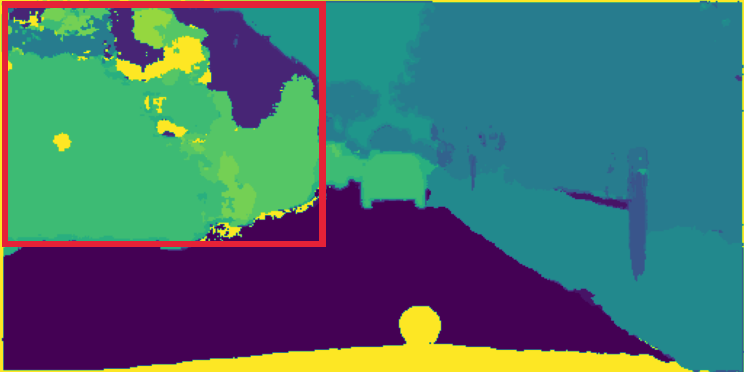}\\
       \rotatebox{\textanglerot}{CBFocal} & \includegraphics[width=\vizsize\textwidth]{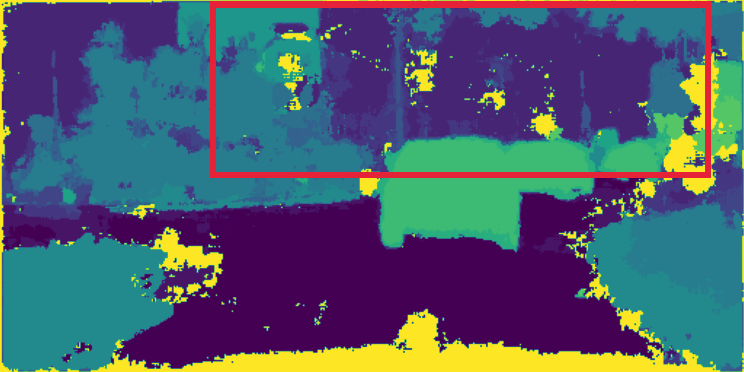} & \includegraphics[width=\vizsize\textwidth]{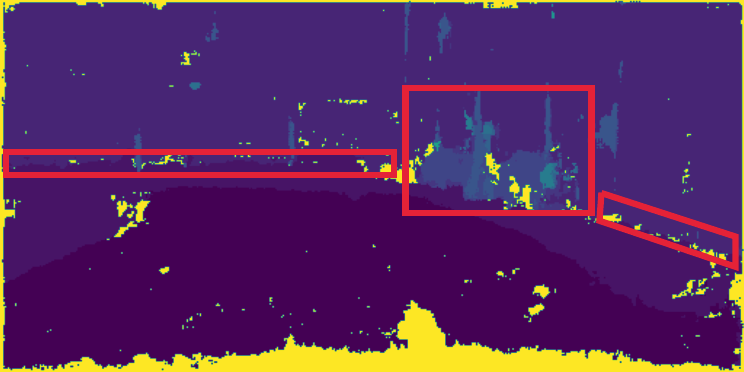}
       & \includegraphics[width=\vizsize\textwidth]{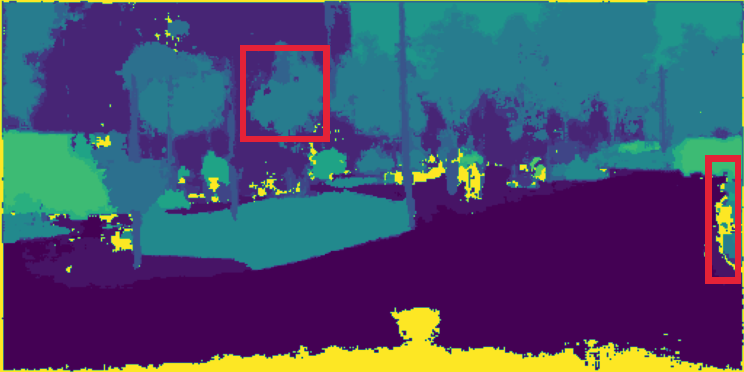} 
       & \includegraphics[width=\vizsize\textwidth]{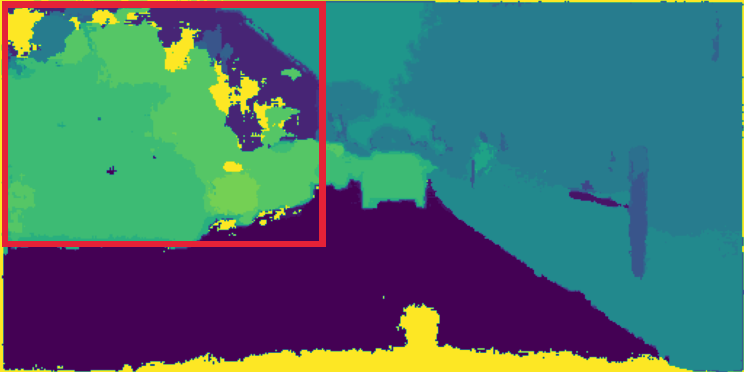}\\
       \rotatebox{\textanglerot}{LDAM} & \includegraphics[width=\vizsize\textwidth]{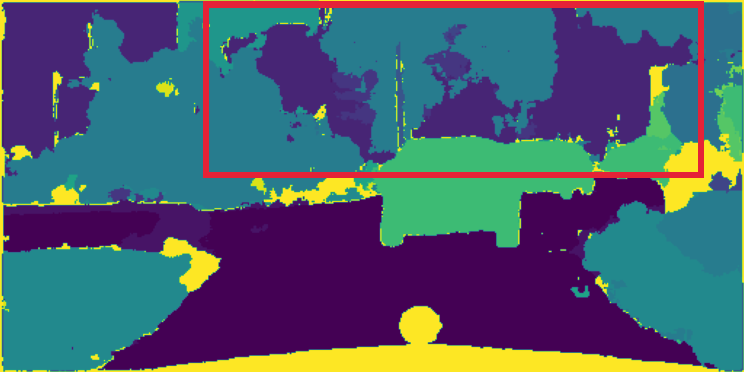} & \includegraphics[width=\vizsize\textwidth]{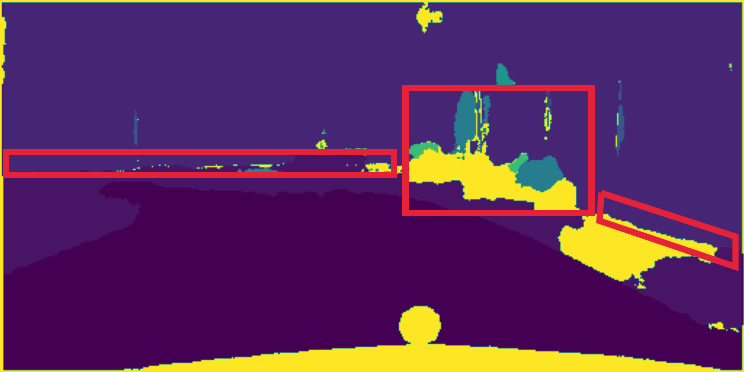}
       & \includegraphics[width=\vizsize\textwidth]{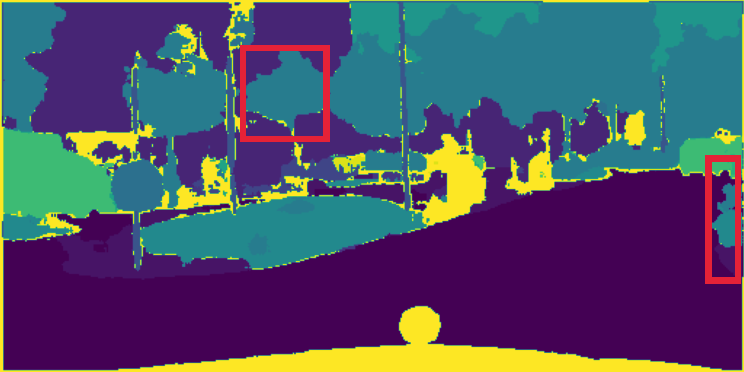} 
       & \includegraphics[width=\vizsize\textwidth]{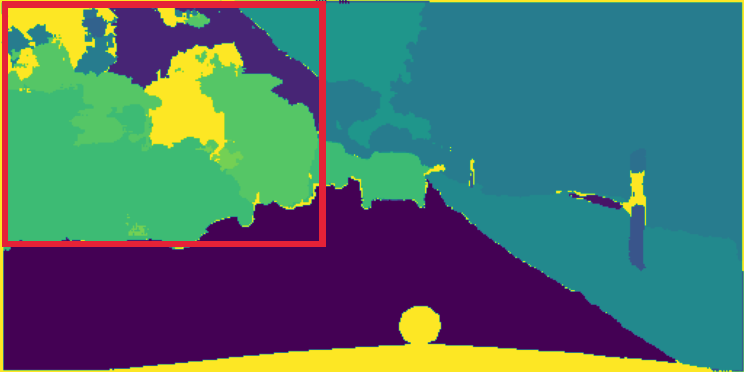}\\
       \rotatebox{\textanglerot}{BLV} & \includegraphics[width=\vizsize\textwidth]{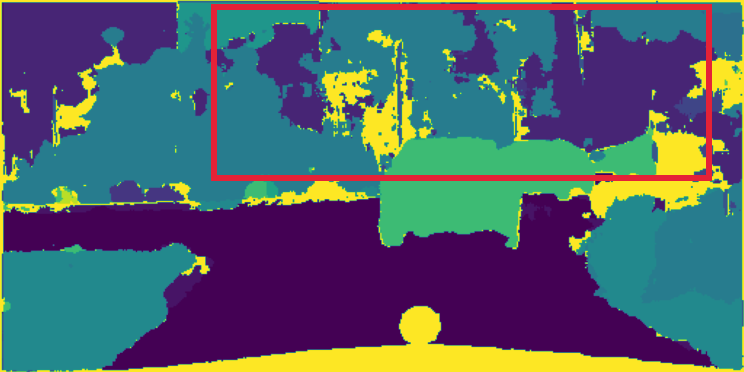} & \includegraphics[width=\vizsize\textwidth]{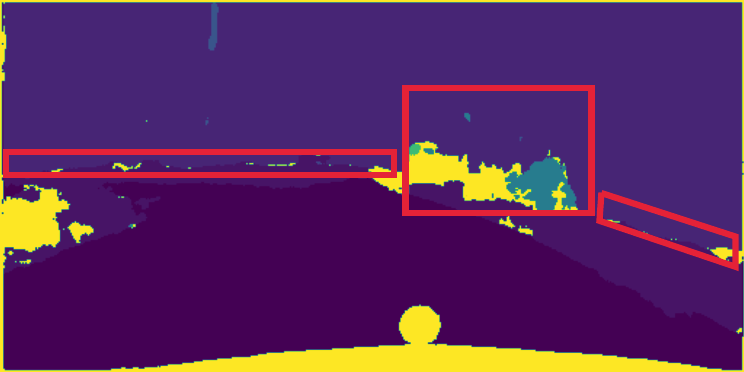}
       & \includegraphics[width=\vizsize\textwidth]{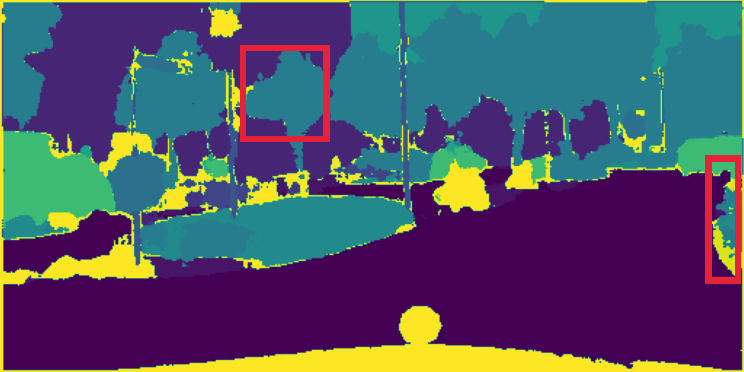} 
       & \includegraphics[width=\vizsize\textwidth]{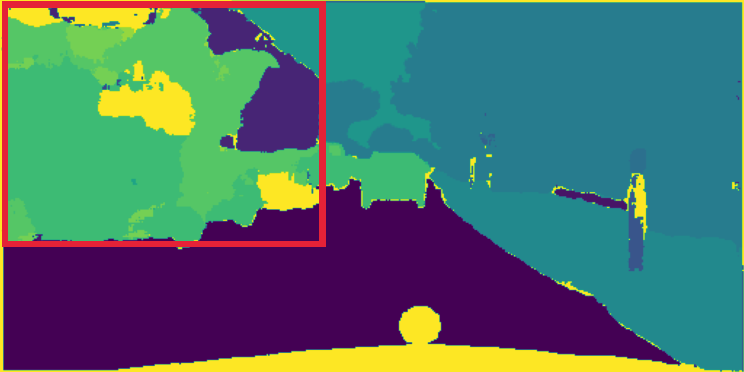}\\
       & Berlin & Munich & Bielefeld & Leverkusen 
    \end{tabular}}
    \caption{Segmentation visualization of models trained by the PAT and other baselines on the CityScapes dataset.}
    \label{fig:perform-city}
\end{figure*}

The metric-based and visual evaluations are shown in Tab. \ref{tab:over} and Fig. \ref{fig:perform-city}. In the second column of Fig. \ref{fig:perform-city} (Munich domain), the model trained by PAT can segment fully the road, sky, and most of the building. Visualization performance in Berlin and Leverkusen (refer to Fig. \ref{fig:perform-city}), are two typical examples. Other SOTAs tend to misclassify the sky and the road, even though the sky and road owned a considerable proportion.

\subsection{Ablation studies}\label{sec:abs}
\subsubsection{Model integration analysis.}\label{sec:abs-model}
To guarantee the PAT's adaptability to different model designs \cite{alexandridis-2022,pan-2021,zang-2021,wang-2021,li-2020,wang-2023}, we experiment with different model structures: UNet\cite{unet}, Attention UNet\cite{attunet} (AttUNet), Nested UNet\cite{nestunet} (UNet++), DeepLabV3\cite{dlv3} (DLV3), and DeepLabV3+\cite{dlv3p} (DLV3+) (refer to Tab. \ref{tab:model}). Tab. \ref{tab:model} shows PAT's superiority on CityScapes, the largest dataset. UNet++ with PAT achieves mIoU and pix acc above 75\% and 85\%, surpassing BLV and LDAM. Class-Balance (CB) excels on OxfordPetIII (3 classes) but struggles with larger datasets.

\begin{table}[!ht]
\centering
\caption{Comparisons between baselines and PAT in different model architecture designs: UNet, Attention UNet, Nested UNet, DLV3, and DLV3+ whose outperforming cases are indicated by green, blue, yellow, red, and purple colors.}
\resizebox{0.4\textwidth}{!}{%
\begin{tabular}{llcccc}
\hline
\multicolumn{1}{c}{\multirow{2}{*}{\textbf{Method}}} &
  \multirow{2}{*}{\textbf{Model}} &
  \multicolumn{2}{c}{\textbf{CityScapes}} &
  \multicolumn{2}{c}{\textbf{NyU}} \\ \cline{3-6} 
\multicolumn{1}{c}{} &
   &
  mIoU$\uparrow$ &
  Pix Acc$\uparrow$ &
  mIoU$\uparrow$ &
  Pix Acc$\uparrow$ \\ \hline
\multirow{5}{*}{\begin{tabular}[c]{@{}l@{}}Focal\\ ($\gamma = 2$)\end{tabular}} &
  \multicolumn{1}{l|}{UNet} &
  73.53 &
  83.69 &
  15.42 &
  51.03 \\
 &
  \multicolumn{1}{l|}{AttUnet} &
  73.93 &
  83.10 &
  16.46 &
  51.32 \\
 &
  \multicolumn{1}{l|}{UNet++} &
  74.21 &
  83.52 &
  16.72 &
  51.41 \\
 &
  \multicolumn{1}{l|}{DLV3} &
  77.91 &
  92.78 &
  22.01 &
  57.64 \\
 &
  \multicolumn{1}{l|}{DLV3+} &
  78.18 &
  93.06 &
  22.35 &
  57.92 \\ \hline
\multirow{5}{*}{CB} &
  \multicolumn{1}{l|}{UNet} &
  \cellcolor[HTML]{79CDCD}\textbf{78.71} &
  83.19 &
  20.18 &
  54.91 \\
 &
  \multicolumn{1}{l|}{AttUnet} &
  74.47 &
  83.76 &
  19.07 &
  54.66 \\
 &
  \multicolumn{1}{l|}{UNet++} &
  74.85 &
  83.66 &
  19.17 &
  53.23 \\
 &
  \multicolumn{1}{l|}{DLV3} &
  77.43 &
  92.52 &
  21.86 &
  57.28 \\
 &
  \multicolumn{1}{l|}{DLV3+} &
  77.77 &
  93.12 &
  22.60 &
  57.96 \\ \hline
\multirow{5}{*}{\begin{tabular}[c]{@{}l@{}}LDAM\\ ($\mu = 0.5$, \\ $s=20$)\end{tabular}} &
  \multicolumn{1}{l|}{UNet} &
  73.42 &
  83.81 &
  18.04 &
  54.33 \\
 &
  \multicolumn{1}{l|}{AttUnet} &
  73.73 &
  83.46 &
  19.81 &
  \cellcolor[HTML]{B0C4DE}\textbf{55.22} \\
 &
  \multicolumn{1}{l|}{UNet++} &
  73.55 &
  84.64 &
  19.04 &
  52.99 \\
 &
  \multicolumn{1}{l|}{DLV3} &
  77.87 &
  92.79 &
  22.28 &
  57.85 \\
 &
  \multicolumn{1}{l|}{DLV3+} &
  78.40 &
  93.55 &
  23.04 &
  58.63 \\ \hline
\multirow{5}{*}{\begin{tabular}[c]{@{}l@{}}BLV\\ (Gaussian,\\ $\sigma = 0.5$)\end{tabular}} &
  \multicolumn{1}{l|}{UNet} &
  74.72 &
  84.74 &
  18.32 &
  \cellcolor[HTML]{79CDCD}\textbf{54.67} \\
 &
  \multicolumn{1}{l|}{AttUnet} &
  74.45 &
  84.56 &
  19.45 &
  54.72 \\
 &
  \multicolumn{1}{l|}{UNet++} &
  74.93 &
  84.86 &
  19.81 &
  \cellcolor[HTML]{FFE4B5}\textbf{54.93} \\
 &
  \multicolumn{1}{l|}{DLV3} &
  78.04 &
  93.04 &
  22.40 &
  57.99 \\
 &
  \multicolumn{1}{l|}{DLV3+} &
  78.42 &
  93.59 &
  23.16 &
  58.71 \\ \hline
\multirow{5}{*}{\begin{tabular}[c]{@{}l@{}}PAT (Ours)\\ ($T = 20$)\end{tabular}} &
  \multicolumn{1}{l|}{UNet} &
  \cellcolor[HTML]{79CDCD}\textbf{74.85} &
  \cellcolor[HTML]{79CDCD}\textbf{85.51} &
  \cellcolor[HTML]{79CDCD}\textbf{21.18} &
  54.22 \\
 &
  \multicolumn{1}{l|}{AttUnet} &
  \cellcolor[HTML]{B0C4DE}\textbf{74.57} &
  \cellcolor[HTML]{B0C4DE}\textbf{85.56} &
  \cellcolor[HTML]{B0C4DE}\textbf{21.44} &
  54.41 \\
 &
  \multicolumn{1}{l|}{UNet++} &
  \cellcolor[HTML]{FFE4B5}\textbf{75.24} &
  \cellcolor[HTML]{FFE4B5}\textbf{85.80} &
  \cellcolor[HTML]{FFE4B5}\textbf{20.66} &
  54.86 \\
 &
  \multicolumn{1}{l|}{DLV3} &
  \cellcolor[HTML]{FFB6C1}\textbf{78.48} &
  \cellcolor[HTML]{FFB6C1}\textbf{93.10} &
  \cellcolor[HTML]{FFB6C1}\textbf{22.66} &
  \cellcolor[HTML]{FFB6C1}\textbf{58.41} \\
 &
  \multicolumn{1}{l|}{DLV3+} &
  \cellcolor[HTML]{D7BDE2}\textbf{78.63} &
  \cellcolor[HTML]{D7BDE2}\textbf{94.01} &
  \cellcolor[HTML]{D7BDE2}\textbf{23.72} &
  \cellcolor[HTML]{D7BDE2}\textbf{59.09} \\ \hline
\end{tabular}
}
\label{tab:model}
\end{table}

\subsubsection{Temperature configurations.}\label{sec:abs-temp}
We perform various experiments of PAT with different temperatures $T\in\{5, 10, 20, 50\}$ (refers to Tab. \ref{tab:temp}). This ablation test analyzes how temperature parameter $T$ affects the performance of the segmentation model. To make a fair comparison, we conduct all experiments with three related datasets as mentioned in Section \ref{sec:evaluation} with three different types of model architecture including SegNet, UNet, and DLV3+.

\begingroup
\setlength{\tabcolsep}{5pt}
\begin{table}[!ht]
\centering
\caption{Quantitative ablation results of various temperatures $T$.}
\resizebox{0.35\textwidth}{!}{%
\begin{tabular}{@{}p{1cm}lccccc@{}}
\toprule
\multirow{2}{*}{\textbf{Model}} &
\multicolumn{1}{c}{\multirow{2}{*}{\textbf{$T$}}} &
\multicolumn{2}{c}{\textbf{CityScapes}} &
\multicolumn{2}{c}{\textbf{NyU}} \\ \cmidrule(l){3-6} 
&
\multicolumn{1}{c}{} &
mIoU$\uparrow$ &
Pix Acc$\uparrow$ &
mIoU$\uparrow$ &
Pix Acc$\uparrow$ \\ \midrule
\multirow{4}{*}{UNet}   & $5$  & 74.17 & 84.82 & 18.23 & 52.05 \\
                        & $10$ & 74.57 & 85.34 & 19.49 & 54.13 \\
                        & $20$ & \textbf{74.85} & 85.51 & 21.18 & 54.22 \\
                        & $50$ & 74.29 & \textbf{85.56} & \textbf{21.32} & \textbf{54.26} \\ \midrule
\multirow{4}{*}{DLV3+}  & $5$  & 78.42 & 93.66 & 23.58 & 58.9  \\
                        & $10$ & 78.61 & 93.76 & 23.65 & 59.04 \\
                        & $20$ & 78.63 & 94.01 & \textbf{23.72} & 59.09 \\
                        & $50$ & \textbf{79.02} & \textbf{94.45} & 23.32 & \textbf{59.59}  \\ \bottomrule
\end{tabular}
}
\label{tab:temp}
\end{table}
\endgroup

\subsubsection{Class-wise performance evaluation}
We investigate the class-wise model performance based mIoU metric on CityScapes Dataset using DeepLabV3+ model architecture (refers to Tab. \ref{tab:cls-wise}). Tab. \ref{tab:cls-wise} suggests that PAT can improve the model performance on both head and tail classes.

\begingroup
\setlength{\tabcolsep}{2pt}
\begin{table*}[!ht]
\centering
\caption{Class-wise experimental evaluation on CityScapes\cite{city} Dataset using DLV3+\cite{dlv3p} based mIoU metric. Note that bold and underlined number indicates the highest and second-highest performance cases.}
\resizebox{0.8\textwidth}{!}{%
\begin{tabular}{@{}llllllllllllllllllllll@{}}
\toprule
\textbf{Method}  & \rotatebox{90}{\textbf{Road}}  & \rotatebox{90}{\textbf{S.Walk}} & \rotatebox{90}{\textbf{Build.}} & \rotatebox{90}{\textbf{Wall}}  & \rotatebox{90}{\textbf{Fence}} & \rotatebox{90}{\textbf{Pole}}  & \rotatebox{90}{\textbf{Light}} & \rotatebox{90}{\textbf{Sign}}  & \rotatebox{90}{\textbf{Veget.}} & \rotatebox{90}{\textbf{Terrain}} & \rotatebox{90}{\textbf{Sky}} & \rotatebox{90}{\textbf{Person}} & \rotatebox{90}{\textbf{Rider}} & \rotatebox{90}{\textbf{Car}} & \rotatebox{90}{\textbf{Truck}} & \rotatebox{90}{\textbf{Bus}} & \rotatebox{90}{\textbf{Train}} & \rotatebox{90}{\textbf{Motor}} & \rotatebox{90}{\textbf{Bike}} & \rotatebox{90}{\textbf{Void}}  & \rotatebox{90}{\textbf{mIoU}}  \\ \midrule
CE      & 99.11 & 78.60  & 92.97  & 63.32 & 59.07 & \underline{61.34} & 64.07 & 73.82 & 94.10  & 52.56   & 95.60 & 78.40  & 56.13 & 94.89 & 84.84 & 85.69 & 82.17 & 70.44 & 68.16 & 97.79 & 77.73 \\
Focal   & \underline{99.48} & 78.34  & 92.89  & \underline{63.62} & \underline{58.91} & 61.09 & 64.45 & 73.78 & 94.44  & 53.04   & 96.09 & 78.51  & 56.55 & 94.66 & 85.32 & 85.52 & 81.90 & 70.21 & 68.34 & 97.71 & 78.18 \\
CB      & 99.33 & 78.67 & 92.95 & 63.58 & 59.24 & 61.27 & 64.02 & 73.59 & 94.05 & 52.70 & 95.14 & 78.45 & 55.96 & 94.79 & 85.03 & 85.55 & 82.25 & 70.71 & 68.97 & 97.69 & 77.77 \\
CBFocal & 99.37 & 78.78 & 93.22 & 63.28 & 59.29 & 61.27 & 64.54 & 74.02 & 94.47 & 52.55 & 96.06 & 78.80 & 56.40 & 95.03 & 85.3 & 85.84 & 82.48 & 70.62 & 68.90 & 98.12 & 78.21  \\
LDAM    & 99.47 & 79.60 & 93.09 & 63.54 & 59.73 & 61.19 & \underline{65.42} & 74.40 & 94.22 & 52.28 & \textbf{96.76} & 78.93 & 56.51 & 95.17 & 85.68 & 85.98 & 82.54 & \textbf{71.30} & 68.40 & 97.84 & 78.40 \\
BLV     & 99.12 & \textbf{79.97} & 93.44 & 63.12 & 59.49 & \textbf{61.86} & \textbf{65.52} & \underline{74.69} & \textbf{95.03} & \underline{53.13} & \underline{96.49} & \textbf{79.95} & \underline{57.36} & \underline{95.31} & 86.39 & \underline{86.35} & \underline{83.57} & \underline{71.14} & 68.83 & 98.71 & 78.42 \\
PAT     & \textbf{99.76} & \underline{79.92} & \textbf{93.89} & \textbf{64.23} & \textbf{60.11} & 61.11 & 65.28 & \textbf{75.63} & \underline{94.95} & \textbf{54.38} & 95.64 & \underline{79.19} & \textbf{57.60} & \textbf{95.80} & \textbf{86.52} & \textbf{86.37} & \textbf{84.32} & 70.23 & \textbf{68.69} & \textbf{99.15} & 78.63      \\ \bottomrule
\end{tabular}
}
\label{tab:cls-wise}
\end{table*}
\endgroup

\subsubsection{Training utilization.}\label{sec:abs-perform}
Owing to the demand of taking full advantage of big data which is not only a large scaled number of samples but also high pixel resolution\cite{survey, dlv3, dlv3p}, a method that is low cost in both computation and memory usage is essential. To investigate the performance of different methods, we use three metrics including the average training time (seconds/epoch), the average memory acquisition, and the average GPU utilization. We calculate these metrics in each epoch and then take the average value once the training is done. 

\begin{figure*}[!ht]
    \centering
    \includegraphics[width=0.8\textwidth]{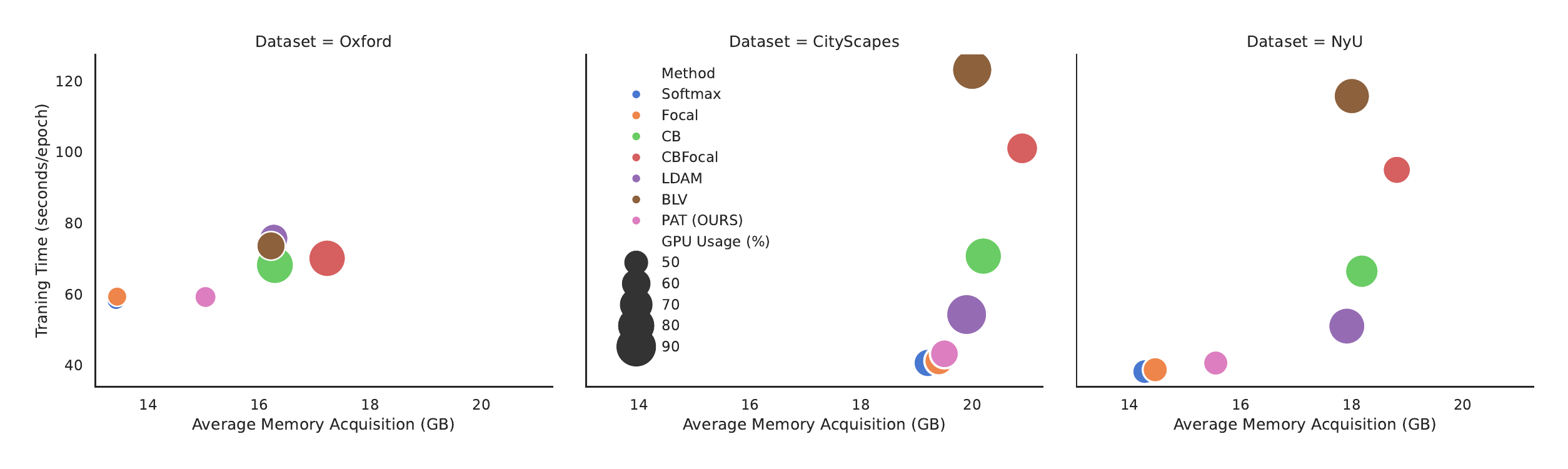}
    \caption{Performance comparison between baselines and our proposed method in three different scenarios containing OxfordPetIII, CityScapes, and NyU. Performance metrics include Training Time (seconds/epoch), Average Memory Acquisition shown in Gigabyte (GB) units, and the GPU Utilization proportion (\%).}
    \label{fig:per}
\end{figure*}

Fig. \ref{fig:per} suggests that PAT (pink circle), which includes the LA and PAT can adapt to a wide range of hardware specifications. While the proposed method acquires roughly 15GB, recent methods (i.e. BLV, LDAM) acquire nearly 17GB and 20GB in the OxfordPetIII and NyU datasets, respectively. In the CityScapes dataset, the proposed method is one of the three lowest GPU-utilized methods, along with the vanilla cross-entropy loss function, which also refers to the lowest time-consumed method.

\section{Conclusion}
We introduce the Pixel-wise Adaptive Training (PAT) technique for long-tailed segmentation. Leveraging class-wise gradient magnitude homogenization and pixel-wise class-specific loss adaptation, our approach alleviates gradient divergence due to label mask size imbalances, and the detrimental effects of rare classes and frequent class forgetting issues. Empirically, on the NyU dataset, PAT achieves a $2.85\%$ increase in mIoU compared to the baseline. Similar improvements are observed in the CityScapes dataset ($2.2\%$ increase) and the OxfordPetIII dataset ($0.09\%$ increase). Furthermore, visualizations reveal that PAT-trained models effectively segment long-tailed rare objects without forgetting well-classified ones.

\bibliographystyle{elsarticle-num}
\bibliography{main.bib}
\end{document}